\documentclass[10pt,twocolumn,letterpaper]{article}

\usepackage{cvpr}              % To produce the CAMERA-READY version
\usepackage{graphicx}
\usepackage{amsmath}
\usepackage{amssymb}
\usepackage{booktabs}

\usepackage{hyperref}
\hypersetup{colorlinks,allcolors=black}

\usepackage[capitalize]{cleveref}
\crefname{section}{Sec.}{Secs.}
\Crefname{section}{Section}{Sections}
\Crefname{table}{Table}{Tables}
\crefname{table}{Tab.}{Tabs.}

\def\cvprPaperID{*****} % *** Enter the CVPR Paper ID here
\def\confName{CVPR}
\def\confYear{2023}

\begin{document}

\title{Technical Report for Argoverse Challenges on Unified Sensor-based Detection, Tracking, and Forecasting}

\author{Zhepeng Wang\textsuperscript{1},  Feng Chen\textsuperscript{1},  
Kanokphan Lertniphonphan\textsuperscript{1},  Siwei Chen\textsuperscript{2}\thanks{Work done as an intern at Lenovo Research.}\label{intern} ,
\\Jinyao Bao\textsuperscript{3}\footnotemark[\value{footnote}],  
Pengfei Zheng\textsuperscript{4}\footnotemark[\value{footnote}],  
Jinbao Zhang\textsuperscript{5}\footnotemark[\value{footnote}],  
Kaer Huang\textsuperscript{1}, 
Tao Zhang\textsuperscript{2}
% \textsuperscript{\Envelope}
\\
\textsuperscript{1}Lenovo Research,\\
\textsuperscript{2}Tsinghua University, 
\textsuperscript{3}Liaoning Petrochemical University,\\
\textsuperscript{4}University of Science and Technology Beijing, \\
\textsuperscript{5}University of Science and Technology of China\\
{\tt\small \{wangzpb, chenfeng13, klertniphonp\}@lenovo.com}
}

\maketitle
%%%%%%%%% ABSTRACT
\begin{abstract}
   This report presents our Le3DE2E solution for unified sensor-based detection, tracking, and forecasting in Argoverse Challenges at CVPR 2023 Workshop on Autonomous Driving (WAD). We propose a unified network that incorporates three tasks, including detection, tracking, and forecasting. This solution adopts a strong Bird's Eye View (BEV) encoder with spatial and temporal fusion and generates unified representations for multi-tasks. The solution was tested in the Argoverse 2 sensor dataset \cite{Argoverse2} to evaluate the detection, tracking, and forecasting of 26 object categories. We achieve 1\textsuperscript{st} place in Detection, Tracking, and Forecasting on the E2E Forecasting track in Argoverse Challenges at CVPR 2023 WAD. 
\end{abstract}

%%%%%%%%% BODY TEXT
\section{Introduction}
\label{sec:intro}

The challenge focuses on evaluating end-to-end perception tasks on detection, tracking, and multi-agent forecasting on Argoverse 2 sensor dataset. The dataset provides track annotations for 26 object categories. For testing, our algorithm needs to be able to detect objects in the current frame and forecast trajectories for the next 3 seconds. The end-to-end task is different from the motion forecasting task since the tracking ground truths are not provided. 

%-------------------------------------------------------------------------
\begin{figure*}[ht]
  \centering
   \includegraphics[width=1.0\linewidth]{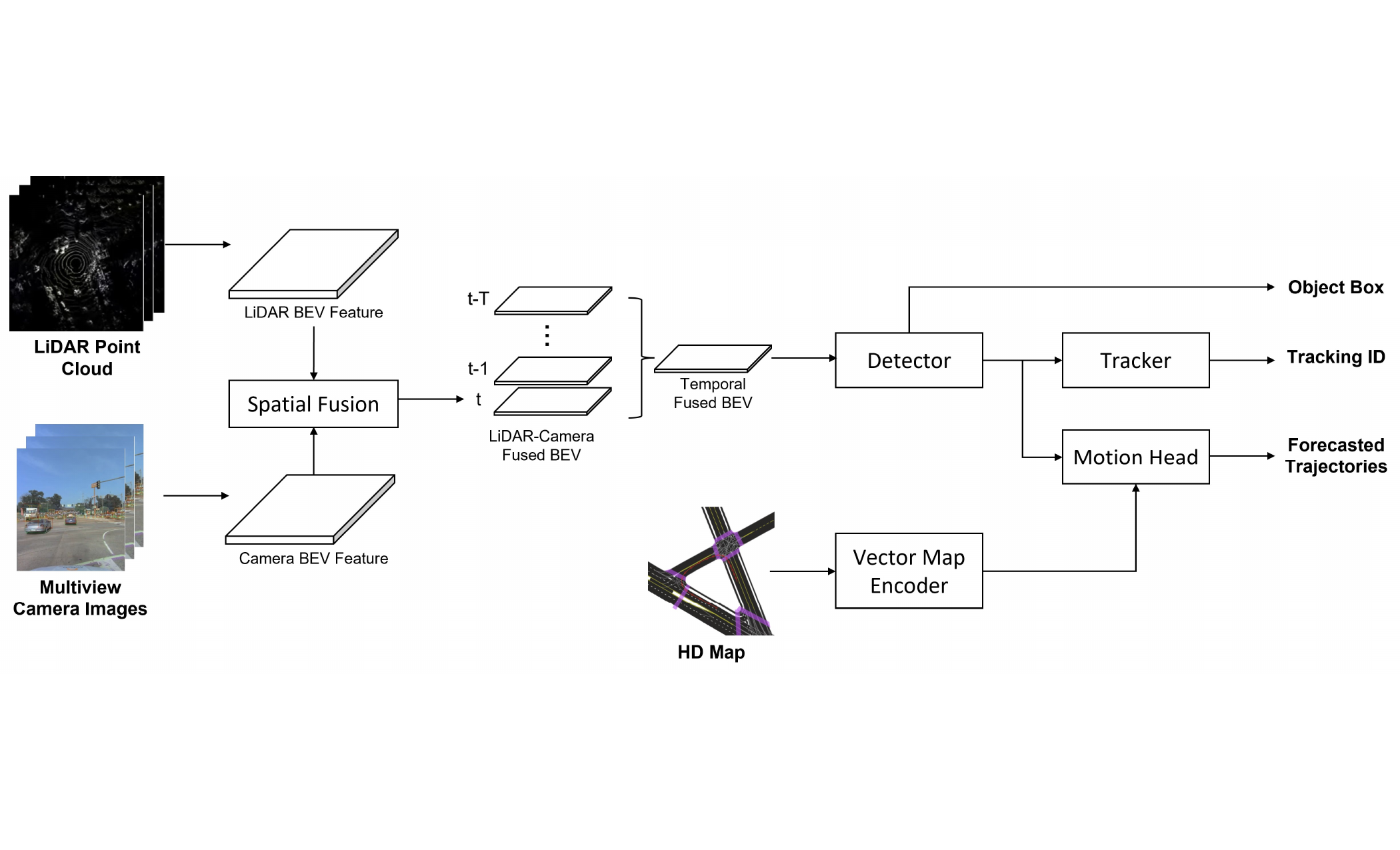}
   \caption{\textbf{System overview.} First, we extract BEV features from LiDAR point cloud and camera images separately. The LiDAR point clouds of the current frame are voxelized and encoded in the BEV feature map by \textbf{LiDAR backbone}. Image features are extracted from synchronized multi-view cameras by an \textbf{image backbone} and are encoded to a camera BEV feature by a transformer-based \textbf{BEV encoder}. Second, \textbf{spatial-fusion} module fuses LiDAR and Camera BEV into a unified BEV representation. The historical frame BEV feature maps are fused with the current frame using a \textbf{temporal encoder}. Third, the spatial-temporal fused BEV is fed into \textbf{Detector} which generates detection bounding boxes. \textbf{Tracker} utilizes object queries from the detector to associate track queries between frames. Furthermore, the \textbf{ Motor Head} forecasts the future trajectories for each agent from Detector. In addition, \textbf{HD Map} is encoded into vectors and interacts with agents to help with motion forecasting.}
   \label{fig:system}
\end{figure*}

\begin{table*}
  \centering
  \begin{tabular}{c c c c}
    \toprule
    Team & mAP\_F($\uparrow$) & ADE($\downarrow$) & FDE($\downarrow$) \\
    \midrule
    dgist$-$cvlab & 45.83 & 4.09 & 4.53 \\
    Host\_4626\_Team & 14.51 & 5.10 & 7.32 \\
    \hline
    \textbf{Le3DE2E (Ours)} & \textbf{46.70} & \textbf{3.22} & \textbf{3.76} \\
    \bottomrule
  \end{tabular}
  \caption{Forecasting Leaderboard on End-to-End Forecasting Challenge}
  \label{tab:forecasting_result}
\end{table*}

\begin{table*}
  \centering
  \begin{tabular}{c c c c}
    \toprule
    Team & HOTA($\uparrow$) & AMOTA($\uparrow$) & MOTA($\uparrow$) \\
    \midrule
    AIDrive (v0) & 44.36 & 17.47 & 32.61 \\
    dgist$-$cvlab & 41.49 & 7.88 & 17.97 \\
    Host\_4626\_Team & 39.98 & 7.10 & 16.21 \\
    \hline
    \textbf{Le3DE2E (Ours)} & \textbf{56.19} & \textbf{19.53} & \textbf{39.34} \\
    \bottomrule
  \end{tabular}
  \caption{Tracking Leaderboard on End-to-End Forecasting Challenge}
  \label{tab:tracking_result}
\end{table*}

\begin{table*}
  \centering
  \begin{tabular}{c c c c c c}
    \toprule
    Team & mCDS($\uparrow$) & mAP($\uparrow$) & mATE($\downarrow$) & mASE($\downarrow$) & mAOE($\downarrow$) \\
    \midrule
    BEV (BEVFusion)	& 0.37 & 0.46 & 0.40 & \textbf{0.30} & 0.50 \\
    Detectors & 0.34 & 0.42 & \textbf{0.39} & \textbf{0.30} & 0.50 \\
    AIDrive (Lv0) & 0.27 & 0.35 & 0.45 & 0.33 & 0.84\\
    Match (lt3d) & 0.21 & 0.26 & 0.43 & 0.33 & 0.50\\
    Host\_75088\_Team (CenterPoint) & 0.14 & 0.18 & 0.49 & 0.34 & 0.72 \\
    zgzxy001 & 0.12 & 0.15 & 0.45 & 0.34 & 0.65\\
    \hline
    \textbf{Le3DE2E (Ours)} & \textbf{0.39} & \textbf{0.48} & 0.41 & 0.31 & \textbf{0.47} \\
    \bottomrule
  \end{tabular}
  \caption{3D Object Detection Leaderboard}
  \label{tab:detection_result}
\end{table*}

%-------------------------------------------------------------------------

\section{Method}
\label{sec:method}

Motivated by UniAD \cite{hu2023_uniad}, we propose an end-to-end framework for detection, tracking, and forecasting. We fuse the BEV features from LiDAR and multi-view cameras as a unified representation for all three downstream tasks. HD map is encoded as vectors to help with motion forecasting. The system overview is shown in figure \ref{fig:system}.
% We adopt TrackFormer and MotionFormer modules for tracking and forecasting from UniAD\cite{hu2023_uniad}, which consists of several transformer decoder-based perception, prediction, and planner modules as

\subsection{BEV Feature}
% \subsubsection{LiDAR BEV Feature}

For LiDAR point cloud, we employ a LIDAR BEV encoder based on SECOND\cite{yan2018second} to generate LIDAR BEV features \(B_l\). For multi-view images, we adopt a spatio-temporal transformer based on BEVFormer\cite{li2022bevformer} to generate BEV features from multi-view cameras  \(B_c\). The camera BEV branch has two modules: the backbone network and the BEV encoder.  

% The LIDAR encoder takes the current and past \(T=4\) frames of the point cloud with a time step of 0.5 s as input. All the past point clouds have been aligned to the current frame.

% \subsubsection{Camera BEV Feature}

% \subsubsection{Spatial and Temporal Fusion}

The BEV features from LiDAR  \(B_l\) and multi-view cameras  \(B_c\) are  fused into one BEV feature by a spatial encoder following BEVFusion\cite{liu2022bevfusion}. The spatial encoder concatenates \(B_l\) and \(B_c\) and then reduces the features dimensions through a convolution layer.  After spatial fusion, historical BEV features are fused with the current frame by the spatial-temporal transformer in BEVformer\cite{li2022bevformer}. The spatial-temporal fused BEV feature is used as a 3D representation and input to downstream heads. 
% or LiDAR-only BEV feature
% We extract 1 camera frame of every 2 frames to synchronize with LiDAR scans. 

\subsection{Detector}

The detector is based on Deformable DETR \cite{Zhu2021_deformable_detr}. The temporal fused BEV features are fed into the decoder as object queries. The Deformable DETR head is used to predict 3D bounding boxes and velocity without Non-Maximum Suppression (NMS). 3D box regression is supervised using the L1 loss. The detection queries capture the agent characteristic by attending to the BEV features.

\subsection{Tracker}

Tracking is initialized by object queries from the detector as the tracking candidate in each frame. While track queries, which are based on MOTR \cite{zeng2021motr}, are used to associate track queries in the current frame and the previous frame. The track queries which are matched with the history frame aggregate temporal information in a self-attention module until the agent disappears in a certain time period.

\subsection{VecterMap Encoder}

% HD Maps play a crucial role in motion forecasting since they provide detailed information about the road, such as lane position and types. 
HD maps are typically represented by vectorized spatial coordinates. To encode the information on lanes and pedestrian crossings, we adopt a vectorized encoding method called VectorNet \cite{Gao2020VectorNetEH}, which operates on the vectorized HD maps to avoid lossy rendering and computationally intensive ConvNet encoding steps. The map elements are encoded by cross-attention layers and are represented as map queries. We generate the position encoding with the center of each vector. The map queries and the position encoding are forwarded to the Motion Head to help with motion forecasting. 

\subsection{Motion Head}

The motion head takes in the agent’s information from the Detection and the map information from the Vector Map Encoder. It then predicts the future trajectories of the agents. The transformer structure has been proven to be effective in motion forecasting tasks in recent years. Thus, we choose MotionFormer from UniAD \cite{hu2023_uniad} as a motion baseline. The motion head is a 3-layer transformer decoder and has BEV queries generated by the BEV encoder, agent queries generated by Detection, and map queries encoded by VectorMap as input. They interact with motion queries and help motion forecasting.

\subsection{Test Time Augmentation and Ensemble}

During inference, we apply Test Time Augmentation (TTA) to further improve the performance. In addition, we use NMS to merge the results of augmented input.  

We use the Weighted Box Fusion (WBF)\cite{solovyev2021weighted} to ensemble multiple models with different training settings to improve detection and forecasting prediction accuracy. For E2E forecasting, we use a two-step ensemble procedure to ensemble not only the detection bounding boxes, but also future trajectories. In step 1, we cluster the detection bounding boxes according to the intersection-over-union (IoU). In step 2, we cluster the forecasting trajectories with L2 distances and adaptively adjust the threshold based on the speed of instances.

%-------------------------------------------------------------------------

\section{Experiments}
\label{sec:experiments}

\subsection{Dataset}

The competition used the Argoverse 2 Sensor Dataset, which consisted of 1000 scenes (750 for training, 150 for validation and 150 for testing) with a total of 4.2 hours of driving data. The total dataset is extracted in the form of 1 TB of data. Each vehicle log has a duration of approximately 15 seconds and includes an average of approximately 150 LiDAR scans with 10 FPS LiDAR frames. The dataset has 7 surrounding cameras with 20 FPS. For the E2E Forecasting track, 1 keyframe is sampled in 2Hz from the training, validation, and testing sets.  

\subsection{Evaluation Metrics}

\noindent\textbf{Detection.} Argoverse\cite{Argoverse2}  proposes a new metric Composite Detection Score (CDS) which simultaneously measures precision, recall, object extent, translation error, and orientation. The mean metrics are computed as an average of 26 different object categories.

\noindent\textbf{Tracking.} HOTA\cite{Luiten2020HOTAAH} is the key metric for the challenge, while AMOTA and MOTA are also important metrics for reference. HOTA explicitly balances the effect of performing accurate detection, association, and localization in a single, unified metric. MOTA combines false positives, missed targets, and identifies switches to compute tracking accuracy. AMOTA, similar to MOTA, is averaged over all recall thresholds to consider the confidence of predicted tracks.

\noindent\textbf{Forecasting.} 
The main evaluation metric is Forecasting mAP (mAP\_F)\cite{Peri2022ForecastingFL}, ADE, and FDE which are averaged over static, and non-linearly moving cohorts. mAP\_F is the key metric for the challenge, which defines a true positive when there is a positive match in both the current timestamp T and the future (final) in the T + N time slot. ADE is an average L2 distance between the best-forecasted trajectory and the ground truth. FDE is an L2 distance between the endpoint of the best-forecasted trajectory and the ground truth. 

\subsection{Implementation Details}

\noindent\textbf{Architecture details.} In the LiDAR branch, the voxel size of the LiDAR encoder is (0.075m, 0.075m, 0.2m) and the point clouds range is limited to [-54m, 54m] x [-54m, 54m] x [-3m, 3m] to adapt the maximum range of E2E forecasting. In LiDAR backbone, we down-sampled voxels to 1/8. For the camera branch, we crop and resize camera images to 976x1440 to save GPU memory. we use the ResNet-101 as a backbone and a 4-layer FPN as a neck to extract features from multiview cameras.    

\noindent\textbf{Training.} We apply a 2-step training procedure. First, we train the detector for 6 epochs. Then we train the entire end-to-end network to optimize the detector, tracker, and motion head simultaneously for 20 epochs. We freeze the LiDAR and image backbones in step 2 to save GPU memory.

The models are trained by AdamW optimizer, with a learning rate of 2e-4, a weight decay of 0.01, and a total batch size of 8 on 8 V100 GPUs. We use cosine annealing to decay the learning rate. We applied CBGS (Class-Balanced Grouping and Sampling) \cite{Zhu2019ClassbalancedGA} to get the expert model for balanced data distribution.

\noindent\textbf{TTA and Ensemble.} For every model, we employ global scaling with [0.95, 1, 1.05] and flipping with respect to the xz-plane and yz-plane for TTA. We trained multiple models with three voxel sizes of [0.05m, 0.075m, 0.1m], with or without CBGS augmentation and with or without camera input. Totally we ensemble 8 models to generate final results.

% We followed the baseline \cite{peri2022towards} for data preparation for training and evaluation. While training detection only, we applied CBGS (Class-balanced Grouping and Sampling) \cite{Zhu2019ClassbalancedGA} to generate a more balanced data distribution.

\subsection{Final Results}

We test our solution on 3 sub-challenges of Detection, Tracking, and Forecasting in the E2E Forecasting track of the Argoverse Challenge. Table \ref{tab:forecasting_result} is the final leaderboard of Forecasting and shows that our solution achieves 46.70 mAP\_F and ranks 1\textsuperscript{st} place in Forecasting. Table \ref{tab:tracking_result} is the final leaderboard of Tracking and shows that our solution achieves 56.19 HOTA and ranks 1\textsuperscript{st} place in Tracking. Table \ref{tab:detection_result} is the final leaderboard of 3D Object Detection and shows that our solution achieves 0.34 CDS and ranks 1\textsuperscript{st} place in Detection.

%-------------------------------------------------------------------------

\section{Conclusion}
\label{sec:conclusion}

We devise a unified framework of detection, tracking, and forecasting for Autonomous Driving. Our solution ranks 1\textsuperscript{st} place in Detection, Tracking, and Forecasting of the E2E Forecasting track in Argoverse Challenges at CVPR 2023 WAD.

%-------------------------------------------------------------------------

%%%%%%%%% REFERENCES
{\small
\bibliographystyle{ieee_fullname}
\bibliography{egbib.bib}
% \bibliography{Main.bbl}

}

\end{document}